\documentclass[runningheads]{llncs}
\usepackage{graphicx}
\usepackage{tikz}
\usepackage{comment}
\usepackage{amsmath,amssymb} % define this before the line numbering.
\usepackage{color}

\usepackage{booktabs}

\usepackage[ruled,vlined]{algorithm2e}
\usepackage{multirow}
\usepackage{mathabx}
\usepackage{xcolor}
\usepackage{algorithmic}

\usepackage{epsfig}
\usepackage{subcaption}
\usepackage{tablefootnote}

\usepackage[pagebackref,breaklinks,colorlinks]{hyperref}

\usepackage{floatrow}
\usepackage{blindtext}
\newfloatcommand{capbtabbox}{table}[][\FBwidth]

\usepackage{caption}

\newcommand{\etal}{\textit{et al. }}

\usepackage[accsupp]{axessibility}  % Improves PDF readability for those with disabilities.

\begin{document}
% % \renewcommand\thelinenumber{\color[rgb]{0.2,0.5,0.8}\normalfont\sffamily\scriptsize\arabic{linenumber}\color[rgb]{0,0,0}}
% % \renewcommand\makeLineNumber {\hss\thelinenumber\ \hspace{6mm} \rlap{\hskip\textwidth\ \hspace{6.5mm}\thelinenumber}}
% % \linenumbers
% \pagestyle{headings}
% \mainmatter
% \def\ECCVSubNumber{5379}  % Insert your submission number here

\title{Image Search with Text Feedback by Additive Attention Compositional Learning} % Replace with your title

% INITIAL SUBMISSION 
%\begin{comment}
% \titlerunning{ECCV-22 submission ID \ECCVSubNumber} 
% \authorrunning{ECCV-22 submission ID \ECCVSubNumber} 
% \author{Anonymous ECCV submission}
% \institute{Paper ID \ECCVSubNumber}
%\end{comment}
%******************

% CAMERA READY SUBMISSION
\titlerunning{Abbreviated paper title}
% If the paper title is too long for the running head, you can set
% an abbreviated paper title here
%
\author{Yuxin Tian\inst{1} \and
Shawn Newsam\inst{1} \and
Kofi Boakye\inst{2}}
\authorrunning{Y. Tian et al.}
% First names are abbreviated in the running head.
% If there are more than two authors, 'et al.' is used.
%
\institute{University of California, Merced 
\email{\{ytian8,snewsam\}@ucmerced.edu} \\ \and
Pinterest \email{kofi@pinterest.com}
}
%******************
\maketitle

\begin{abstract}
Effective image retrieval with text feedback stands to impact a range of real-world applications, such as e-commerce. Given a source image and text feedback that describes the desired modifications to that image, the goal is to retrieve the target images that resemble the source yet satisfy the given modifications by composing a multi-modal (image-text) query. 
We propose a novel solution to this problem, Additive Attention Compositional Learning (AACL), that uses a multi-modal transformer-based architecture and effectively models the image-text contexts. Specifically, we propose a novel image-text composition module based on additive attention that can be seamlessly plugged into deep neural networks. We also introduce a new challenging benchmark derived from the Shopping100k dataset. AACL is evaluated on three large-scale datasets (FashionIQ, Fashion200k, and Shopping100k), each with strong baselines. Extensive experiments show that AACL achieves new state-of-the-art results on all three datasets.
% \keywords{We would like to encourage you to list your keywords within
% the abstract section}
\end{abstract}

%%%%%%%%% BODY TEXT
\section{Introduction}
\label{sec:intro}
Image retrieval is a fundamental task in computer vision and serves as the cornerstone for a wide range of applications such as fashion retrieval~\cite{2020amazonretrieval,2020kdd}, geolocalization~\cite{7299135,9093403}, and face recognition~\cite{sun2014deep}. There are several ways to formulate the search query such as keywords~\cite{Ak_2018_CVPR,8100135}, a query image~\cite{Yang_2021_ICCV,Warburg_2021_ICCV}, or even a sketch~\cite{Ghosh_2019_ICCV,Yelamarthi_2018_ECCV,7780462,Bhunia_2021_CVPR,Bhunia_2020_CVPR,Sain_2021_CVPR}. However, a core challenge in traditional image retrieval is that it is difficult for the user to refine the retrieved items based on their intentions. A range of approaches to incorporate user feedback to refine the retrieved images have been explored. Combining natural language feedback with a query image is a particularly promising framework since it provides a natural and flexible way for users to convey the image modifications that they have in mind. 

In this work, we investigate image retrieval with text feedback where the goal is to retrieve images that are similar to a query image but incorporate the modifications described by the text. 
Such multi-modal and complementary input provides users with a powerful and intuitive visual search experience. However, as a multi-modal learning problem, it requires the synergistic understanding of both visual and linguistic content which can be a challenge. While image search with text feedback lies at the intersection of vision and language analysis, it differs from other extensively studied vision-and-language tasks, such as image-text matching~\cite{2019matching,Lee_2018_ECCV,Zhou_2020_CVPR,jain2021mural}, image captioning~\cite{radford2021learning,Pan_2020_CVPR,Cornia_2020_CVPR}, and visual question answering~\cite{balanced_vqa_v2,Jiang_2020_CVPR,Chen_2020_CVPR,Cadene_2019_CVPR}. 
{This difference stems from the significant challenge of learning a \emph{composite representation} that jointly captures the \emph{visual} content of the query image and the \emph{linguistic} information in the accompanying text to match the target image of interest.}

 \begin{figure}[!t]
    \centering
    \setlength{\belowcaptionskip}{-100pt} 

    \includegraphics[width=0.7\linewidth]{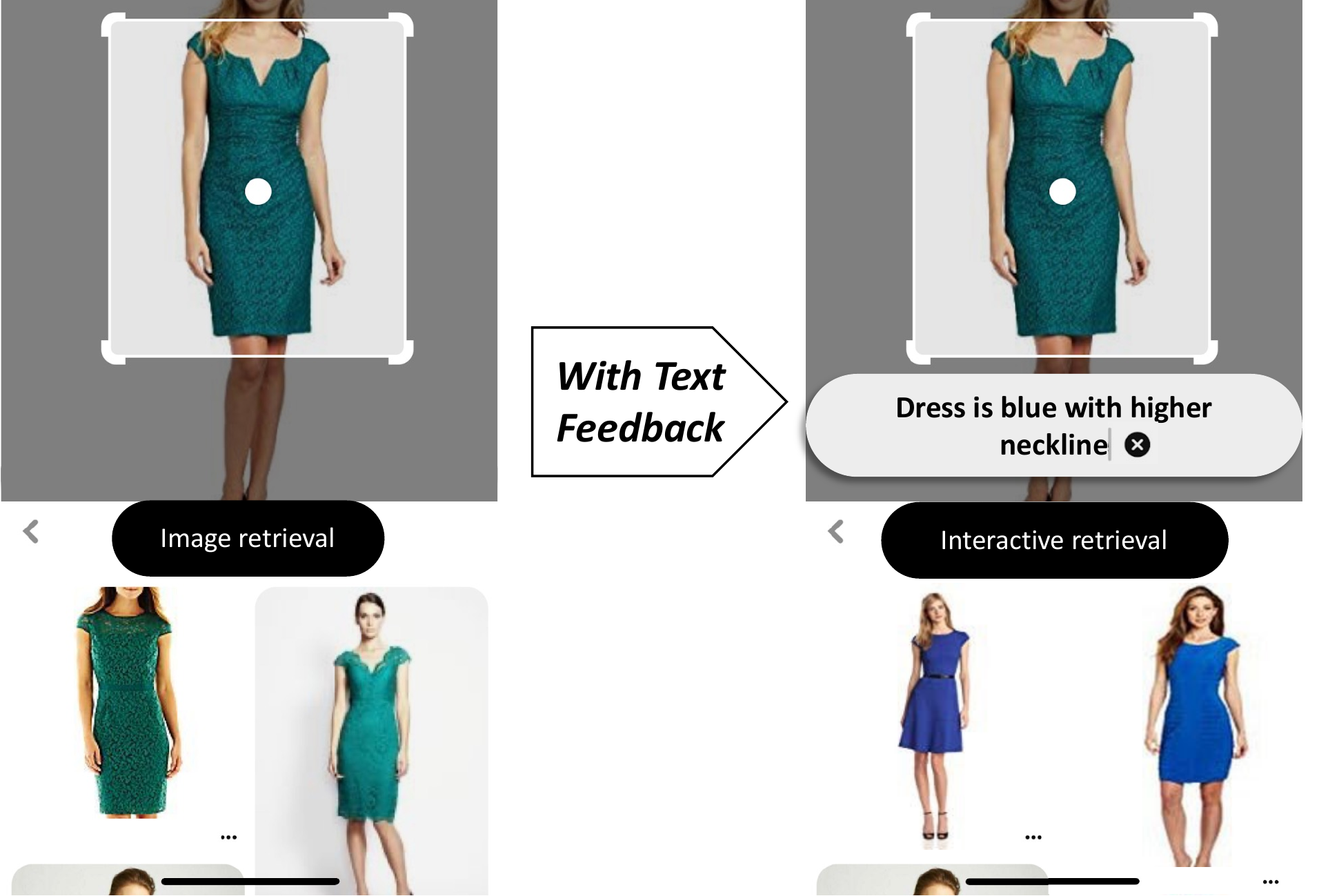}
    \caption{We consider the task of retrieving new images that resemble the reference image while changing certain aspects as specified by text. {Best viewed in color.}}
\vspace{-5pt}
    \label{fig:figure1}
\end{figure}

A fundamental challenge in image-text compositonal learning is characterizing global concepts from the query image and text representation simultaneously. For instance, when the text describes a modification to the color and neckline of a dress in a query image, the composition module should capture the concept of transforming the color and neckline, but it should also preserve the other visual concepts such as the trim, and material of the dress (Figure~\ref{fig:figure1}). Another challenge is how to \emph{selectively} modify the query image representation using the captured contextual information so that it is close to the target image representation in the latent space.

We propose a novel transformer-based Additive Attention Compositional Learning (AACL) model to address these challenges. The key idea is that we learn a contextual vector from the joint visiolinguistic representation. AACL then selectively modifies the query image tokens using the global context vector such that the composite features preserve the  visual content of the image that should not be changed while transforming the relevant content according to the accompanying text.

We empirically compare our AACL approach with the state-of-the-art (SOTA) methods for visual search with text feedback on three large-scale fashion datasets: FashionIQ~\cite{fashioniq}, Fashion200k~\cite{fashion200k}, and a new challenging benchmark derived from Shopping100k~\cite{shop100k}. We show that our proposed compositional learning method outperforms existing methods on all three datasets.

We make the following fundamental contributions:
\begin{itemize}
    \vspace{-5pt}
    \item We propose a novel multi-modal additive attention layer capable of learning a global context vector which is used to selectively modify the image representation in an efficient way. 
    \item We develop a fully transformer-based model for the challenging task of visual search with text feedback and demonstrate that it achieves state-of-the-art performance through extensive experiments on several large-scale fashion datasets. 
     \item We create a new image-text retrieval dataset derived from Shopping100k. This new dataset features a wider range of fashion categories and attributes, resulting in an additional challenging benchmark for the research community.

\end{itemize}

%----------------------------------------------------------------------------------------------------------------------

\section{Related Work}

\subsection{Image Retrieval with Text Feedback}
Image retrieval with text feedback has been of interest to the computer vision research community for some time and a number of efforts (e.g., \cite{appalaraju2021docformer,vilBERT,tirg}) have investigated effective ways to combine image and text representations. The text feedback can be provided in various ways, including absolute attributes (e.g., ``red'')~\cite{Ak_2018_CVPR,8100135,fashion200k}, simple relative attributes (e.g., ``more red'')~\cite{2011attr,2015attr,2019attr}, or full natural language phrases~\cite{tirg,composeAE,j2020sac,chen2020image,maaf,shin2021rtic,kim:2021:AAAI}. 
Natural language is the preferred method of interaction between humans and computers in contemporary search engines. For image search in particular, it allows a user to convey detailed and precise specifications or modifications in a very natural way. We therefore focus on query-based image search with accompanying natural language phrases. 
%%%
%%%Image retrieval with text feedback has been of interest to the computer vision community for some time. Several efforts (e.g., \cite{appalaraju2021docformer, vilBERT,tirg}) have sought to investigate effective ways to combine image and text representations. 
%In general, the 
%%%The text feedback can be given in various formats, including relative attribute\cite{2011attr, 2015attr, 2019attr}, absolute attribute \cite{Ak_2018_CVPR, 8100135, 2017han}, or natural language \cite{tirg, composeAE, j2020sac, chen2020image,maaf,shin2021rtic, kim:2021:AAAI}. 
%%%Natural language is considered the most pervasive method of interaction between human and computer in contemporary search engines.  In addition, it naturally serves to convey concrete information that elaborates a user’s intricate specifications for image search. Consequently, in this work, we investigate image search with natural language queries. 

Previous methods~\cite{composeAE,chen2020learning,kim:2021:AAAI,maaf,shin2021rtic} for image retrieval with text feedback rely heavily on convolution to aggregate features. In contrast, ours is the \emph{first approach to efficiently learn features globally via attention}. Previous works have also relied on complicated hierarchical feature aggregation~\cite{chen2020image,j2020sac}, multiple forms of text feedback~\cite{chen2020image,composeAE}, or multiple loss functions ~\cite{chen2020image,j2020sac,composeAE}. The winning solutions~\cite{kim:2021:AAAI,kim2021cycled,shin2020fashioniq} for the FashionIQ 2020 challenge---an interactive image retrieval challenge---employed common performance boosting techniques such as careful hyperparameter tuning and model ensembles to improve the results. In contrast, AACL focuses on the \emph{design of the image-text composition module} and achieves state-of-the-art performance via feature fusion in one step, which is more efficient and easier to adapt to other frameworks.  

\subsection{Image-Text Composition}\label{sec:relatedwork}

While there has been much effort and different kinds of methods proposed to achieve the top scores on benchmarks involving image and text, relatively few have focused on the {image-text composition module} itself.
In~\cite{kim2016multimodal}, the authors propose a multi-modal residual network (MRN) that learns representations by fusing visual and textual features through element-wise multiplication and residual learning. FiLM~\cite{perez2018film} utilizes a linear modulation component in which text information modifies the image representation via a feature-wise affine transformation. 
Vo \etal  proposed TIRG~\cite{tirg}, which uses a gating mechanism to determine the channels of the image representation that should be modified by the conditioning text. In ComposeAE~\cite{composeAE}, a complex embedding space that semantically ties the representations from text and image modalities is designed. Recently, MAAF~\cite{maaf} improved multi-modal image search via a Modality-Agnostic Attention Fusion model. This model uses a dot product attention mechanism as found in the standard transformer architecture. Additionally, resolution-wise pooling is proposed to aggregate fine-grained features from a ResNet~\cite{resnet} CNN. RTIC~\cite{shin2021rtic} consists of a residual text and image composer to encode the errors between the source and target images in the latent space and includes a graph convolutional network for regularization. Our work differs from these composition modules in that we utilize a novel image and text composition module via additive attention~\cite{bahdanau2014neural,2015additive} to model global contexts. Furthermore, we use an element-wise product to model the interaction between the global context and each input token, which both greatly reduces the computational cost and effectively captures the contextual information~\cite{kim2016multimodal,kim:2021:AAAI,wu2021fastformer}.
\vspace{-10pt}
\subsection{Attention Mechanism}
\vspace{-5pt}
 The concept of attention has gained popularity recently in neural networks as it allows the models to learn representations from different modalities~\cite{kim2016multimodal,2019Captioning,maaf,chen2020image,appalaraju2021docformer}. The two most commonly used attention functions are additive~\cite{bahdanau2014neural}, and dot-product (multiplicative) attention~\cite{vaswani2017attention}. Dot-product attention has a drawback, however, in that it has to attend to all the tokens on the source side for each target token, which is expensive and can potentially be impractical for longer sequences. Additive attention has been shown experimentally to achieve higher accuracy than multiplicative attention in some scenarios~\cite{2015additive,wu2021fastformer}. Inspired by this, we propose an \emph{additive attention composition module} for feature fusion. 
%  \textcolor{blue}{To the best of our knowledge, we are the first to apply additive attention to multi-modal feature fusion.}
%  In addition, our additive attention adopts the concept of a single global context vector as in [35] and apply it to the architecture in [51]. The result is a simplified and more efficient architecture that, based on our preliminary experiments, performs comparably.}
%----------------------------------------------------------------------------------------------------------------------

\vspace{-5pt}
\section{Method}
\begin{figure*}[!t]
    \centering
    \setlength{\abovecaptionskip}{-0.1cm}
    \setlength{\belowcaptionskip}{-0.1cm} 
    \includegraphics[width=\linewidth]{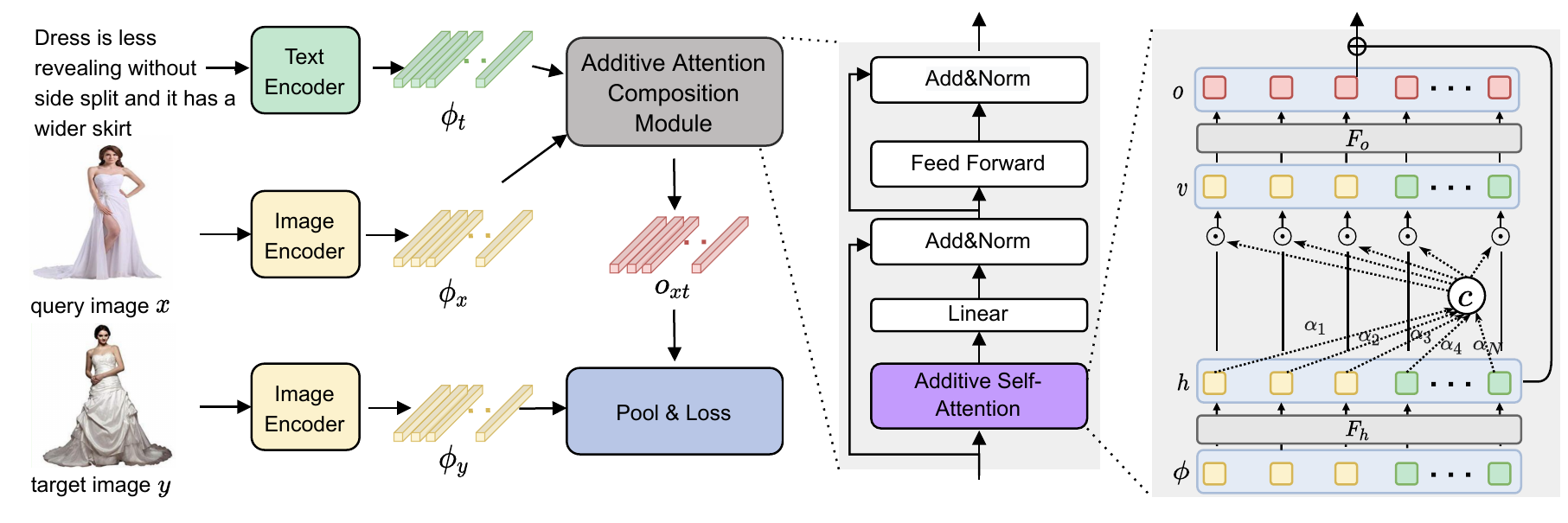}
    % \vspace{-20pt}
    \caption{Overview of our Additive Attention Compositional Learning framework. Given a pair of query image and text as input, our goal is to learn a composite representation that aligns to the target image representation. AACL contains three major components: an image encoder (Sec.~\ref{sec:image_encoder}), a text encoder (Sec.~\ref{sec:text_encoder}), and an Additive Attention Composition Module (Sec~\ref{sec:aacl})
    that can be plugged into different models for feature fusion. ``$\odot$" represents Hadamard product.}
% \vspace{-5pt}
    \label{fig:diagram}
\end{figure*}

\vspace{-5pt}
Figure~\ref{fig:diagram} presents the overall architecture of our Additive Attention Compositional Learning (AACL) framework. Given a source image $x$ and text feedback $t$ as the input query, the goal of AACL is to learn a composite representation $o_{xt}$ that can be used to retrieve relevant images $y$ 
% [NEED DIFFERENT SYMBOL]
from a target database. AACL contains three key components: (1) an image encoder for visual semantic representation learning, (2) a text encoder for natural language representation learning, and (3) an additive attention composition module that modifies the source image representation according to the text representation. 
% All components are jointly optimized in an end-to-end manner via one-turn [NOT SURE WHAT ``one-turn" MEANS HERE] feature composition. 
% single stage late fusion, not multiple
In contrast to other approaches that use multiple stages of feature composition and matching (e.g., \cite{chen2020image}), AACL does this in one stage using the final output of the image and text encoders.
%%%
%%%Figure~\ref{fig:diagram} presents the overall architecture of our Additive Attention Compositional Learning (AACL) framework. Given a source image $x$ and text feedback $t$ as input query, the ultimate aim of AACL is to learn a composite representation $o_{xt}$ that can be used to retrieve target images $y$ 
% [NEED DIFFERENT SYMBOL]
%%%from a target database. AACL contains three components: (1) an image encoder for visual semantic representation learning, (2) a text encoder for natural language representation learning, and (3) an additive attention composition transformer that modifies the source image representation according to the text representation. 
% All components are jointly optimized in an end-to-end manner via one-turn [NOT SURE WHAT ``one-turn" MEANS HERE] feature composition. 
% single stage late fusion, not multiple
%%%In contrast to other approaches that use multiple stages of feature composition and matching (e.g., \cite{chen2020image}), AACL does this in one stage using the final output of the image and text encoders.

In the following, we first provide an overview of the two encoders in Section~\ref{sec:encoders}. We then detail our novel composition module in Section~\ref{sec:aacl} and our model optimization in Section~\ref{sec:loss}.
% Finally, we describe our interactive retrieval pipeline in Section~\ref{sec:pipeline}.

%%%We start with an overview of the two encoders in Section~\ref{sec:encoders}, then elaborate our key composition module and model optimization in Sections~\ref{sec:aacl} and~\ref{sec:loss}, respectively. Finally, the interactive retrieval pipeline is introduced in Section~\ref{sec:pipeline}.

\subsection{Image and text representation} \label{sec:encoders}
\noindent\textbf{Image Representation:} \label{sec:image_encoder}
We employ a Swin Transformer~\cite{liu2021Swin} to derive a discriminative representation of the visual content of an image. As a transformer inherently learns visual concepts of increasing abstraction in a compositional, hierarchical order, we conjecture that image features from the final layer may not fully capture the visual information of the lower levels. We thus concatenate image tokens extracted from the final (Stage 4) and penultimate (Stage 3) layers of the Swin Transformer. Unless otherwise specified, our model uses these $49+49 = 98$ image tokens for multi-level image understanding. A learned linear projection maps each image token to $d$ dimensions so that the final image representation is $\phi_x \in \mathbb{R}^{98 \times d}$.

%%%To encode the visual content into discriminative representations, we employ the Swin transformer~\cite{liu2021Swin}. As a transformer inherently learns visual concepts of increasing abstraction in a compositional, hierarchical order, we conjecture that image features from the final layer may not fully capture the visual information of lower levels. Thus, we concatenate image tokens extracted from the final (Stage-4) and penultimate (Stage-3) layers of the Swin transformer. Unless otherwise specified, our model uses these $49+49 = 98$ image tokens for multi-level image understanding. A learned linear projection is applied to map each original image token to the final image representation: $\phi_x \in \mathbb{R}^{98 \times d}$.

\noindent\textbf{Text Representation:} \label{sec:text_encoder}
The DistilBERT language representation model~\cite{sanh2019distilbert} is used to encode the semantics of the accompanying text. DistilBERT naturally yields $m$ tokens for the input words, namely the hidden states of the last layer of the model. We concatenate these tokens to form the final text representation $\phi_{t} \in \mathbb{R}^{m \times d}$.

%%%The DistilBERT language representation model~\cite{sanh2019distilbert} is utilized as the text encoder to represent the text semantics. DistilBERT naturally yields $m$ tokens for input words, namely the hidden states of the last layer. $\phi_{t} \in \mathbb{R}^{m \times d}$ is the concatenation of these tokens. 
% for each input word token, namely the hidden states of the last layer, where $m$ is the max length of text word tokens [CONFUSING SENTENCE]. 
% Two sent, phi of t is tokens concat ... 

\subsection{Additive Attention Composition Module}
\label{sec:aacl}
% [HOW DO YOU BOTH TRANSFORM AND PRESERVE AT THE SAME TIME?] 
In order to jointly represent the image and text components of the query, we seek to transform the visual features conditioned on language semantics.
% Besides, the additive attention has experimentally shown to achieve higher accuracy than multiplicative attention and easier to train~\cite{2015additive} [THIS IS A RISKY CLAIM]. 
To accomplish this, we propose an \emph{additive attention composition module} for feature fusion. This module consists of multiple composition blocks that each employ additive self-attention to learn a context vector which then selectively modifies the joint visiolinguistic representation. The final output of these blocks yields a modified image representation that is meant to faithfully capture the input image and text information.

\noindent\textbf{Visiolinguistic Representation:}
In order to obtain the input representation for our first composition block, the image tokens $\phi_{x}$ and text tokens $\phi_{t}$ are concatenated to obtain the visiolinguistic representation $\phi=\left[\phi_{x}, \phi_{t}\right]$. The final representation is denoted as $\phi \in \mathbb{R}^{N \times d}$, where $N$ is the combined count of image and text tokens. 

\noindent\textbf{Composition Block:}
Following the standard transformer architecture~\cite{vaswani2017attention}, the additive attention composition module is composed of a stack of $L$ identical blocks with multiple heads. Different attention heads use the same formulation but different parameters, which allows the model to jointly attend to information from different representation subspaces at different positions. Each block has an additive self-attention layer followed by a linear layer and a feed-forward neural network. We also employ a residual connection and layer normalization after these linear and feed-forward components.

\noindent\textbf{Additive Self-Attention Layer:}
In order to discover the latent relationships essential for learning the transformation, we use the additive attention mechanism to learn a context vector $c$, then selectively suppress and highlight the representations from each token.
Similar to~\cite{wu2021fastformer}, we first use a linear transformation layer to transform the input sequence into the hidden states: $h = \mathcal{F}_{h}{\left(\phi_i \right)}, i\in N$. The context vector $c$ that is learned to modify each token is generated as a weighted sum of these tokens $h_i$: 
\begin{equation}
c=\sum_{i=1}^{{N}} \alpha_{i} h_{i},
\end{equation}
The weight $\alpha_i$ of each token $h_i$ is computed by
\begin{equation}\label{eq:alpha}
\alpha_{i}=\frac{\exp \left(\mathbf{w}_{h}^{T} \mathbf{h}_{i}/ \sqrt{d}\right)}{\sum_{j=1}^{N} \exp \left(\mathbf{w}_{h}^{T} \mathbf{h}_{j}/ \sqrt{d}\right)}.
\end{equation}
where $\mathbf{w}_h \in \mathbb{R}^{d}$ is learned during the training process, and $\mathbf{w}_{h}^{T} \mathbf{h}_{j}$ scores how much each input token contributes to the global context.

Next, to selectively suppress and highlight the visual content in $h$, a Hadamard product is introduced to reuse the global contextual information, which is motivated by its effectiveness in modeling the nonlinear relationship between two vectors~\cite{2017product,wu2021fastformer,hu2018senet}. It is formulated as $v_i = c \odot h_i$. Another linear transformation layer $\mathcal{F}_{o}$ is applied to each token $v_i$ to learn its hidden representation. To form the final output of the additive attention layer, we add the hidden states $h_i$ that capture relevant source-side information to the transformed latent features. The final output of the additive self attention layer is: 
\begin{equation}
{o}_{{i}}=h_i + \mathcal{F}_{o}\left(c \odot h_i\right)
\end{equation}

% \vspace{-20pt}
\subsection{Deep Metric Learning}
\label{sec:loss}
Our objective during training is to push the ``modified'' image representation $\phi_{xt}$ and the target image representations $\phi_y$ closer, while pulling apart the representations of dissimilar images. A batch-based classification loss as in~\cite{tirg,composeAE,maaf} is used to train the model as early experiments showed that the triplet loss performs worse for the Recall@k metric. Each batch is constructed from $N$ pairs of a query (image and text) and its corresponding target image.
\begin{equation}
L=\frac{1}{B} \sum_{i=1}^{B}-\log \left\{\frac{\exp \left\{\kappa\left(\phi_{y}, \phi_{xt}\right)\right\}}{\sum_{j=1}^{B} \exp \left\{\kappa\left(\phi_{y}, \phi_{xt}\right)\right\}}\right\}
\end{equation}
where $B$ is the batch size and $\kappa$ is a similarity kernel that is implemented as the dot product in our experiments.

%----------------------------------------------------------------------------------------------------------------------
\vspace{-10pt}
\section{Experiments}
\vspace{-10pt}
\subsection{Experimental Setup}

\noindent\textbf{Datasets:} 
We evaluate our model on three datasets---FashionIQ, Fashion200k and our modified version of Shopping100k---in order to validate its ability to generalize to a variety of natural language expressions. We provide details of these datasets in Sections~\ref{fashionIQ}, \ref{fashion200k}, and \ref{shopping100k}, respectively.

\noindent\textbf{Implementation Details:}
We use the PyTorch deep learning framework to conduct all our experiments.  The Swin Transformer~\cite{liu2021Swin} is used as the backbone for the image encoder. The transformer model is initialized using weights first pre-trained on ImageNet-22K and then fine-tuned on ImageNet-1K~\cite{imagenet_cvpr09}.

We extract sequences of 1024-dimensional tokens from Stages 3 and 4 of the model and then project the tokens to $d$ dimensions, which for our experiments is 768. 
We learn the text embedding using a pre-trained DistilBERT model~\cite{sanh2019distilbert}, which yields a 768-dimensional token for each input word. The original BERT model is pre-trained on BooksCorpus (800M words) and English Wikipedia (2,500M words)~\cite{devlin2018bert}. We employ 3 additive attention composition blocks and 8 parallel attention heads for each block. For training, we use SGD optimization with a learning rate of 0.035. We train all models using 4 GPUs with a batch size of 32 per GPU. For FashionIQ, we employ a learning rate decay of 0.1 every 10 epochs for 60 epochs. For Fashion200k and our modified Shopping100k, we use the same decay value but every 30 epochs with a total of 100 epochs. We report the average and standard deviation of five trials for all our experiments to obtain more meaningful results.

\noindent\textbf{Evaluation Metric:}
For evaluation we adopt Recall@K (denoted as R@K for short), a standard metric in retrieval.

\begin{table*}[!t]
\centering
\resizebox{1.0\linewidth}{!}{
\begin{tabular}{ll|ll|ll|ll|l}

\multirow{2}{*}{Model}              & \multicolumn{2}{|c|}{Shirt}     & \multicolumn{2}{c|}{Dress}    & \multicolumn{2}{c}{Toptee}     & \multicolumn{1}{|c}{Average}                      \\
\cline{2-8} 
               & \multicolumn{1}{|c}{R@10}       & \multicolumn{1}{c|}{R@50}     & \multicolumn{1}{c}{R@10} & \multicolumn{1}{c|}{R@50} & \multicolumn{1}{c}{R@10} & \multicolumn{1}{c}{R@50} & \multicolumn{1}{|c}{(R@10 + R@50)/2}\\ 
\toprule
MRN~\cite{kim2016multimodal} & \multicolumn{1}{|c}{15.88}    & \multicolumn{1}{c|}{34.33}   & \multicolumn{1}{c}{12.32}    &  \multicolumn{1}{c|}{32.18}     &  \multicolumn{1}{c}{18.11}   &  \multicolumn{1}{c|}{36.33}  & \multicolumn{1}{c}{24.86}     \\
FiLM~\cite{perez2018film}      & \multicolumn{1}{|c}{15.04}  & \multicolumn{1}{c|}{34.09}    & \multicolumn{1}{c}{14.23}     &  \multicolumn{1}{c|}{33.34}     &  \multicolumn{1}{c}{17.30}   &  \multicolumn{1}{c|}{37.68} & \multicolumn{1}{c}{25.28}   \\
TIRG~\cite{tirg}           & \multicolumn{1}{|c}{16.12}  & \multicolumn{1}{c|}{37.69}   & \multicolumn{1}{c}{19.15}          &  \multicolumn{1}{c|}{43.01}     &  \multicolumn{1}{c}{21.21}   &  \multicolumn{1}{c|}{47.08}   & \multicolumn{1}{c}{30.71}  \\
ComposeAE~\cite{composeAE}     & \multicolumn{1}{|c}{9.96}      & \multicolumn{1}{c|}{25.14}           & \multicolumn{1}{c}{10.77}           &  \multicolumn{1}{c|}{28.29}     &  \multicolumn{1}{c}{12.74}   &  \multicolumn{1}{c|}{30.79}   & \multicolumn{1}{c}{19.61}    \\
% VAL~\cite{chen2020image}  & \multicolumn{1}{|c}{21.03}     &  \multicolumn{1}{c|}{43.44}   & \multicolumn{1}{c}{21.12}           & \multicolumn{1}{c|}{42.19} &  \multicolumn{1}{c}{25.64}   &  \multicolumn{1}{c|}{49.49}  & \multicolumn{1}{c}{33.82}    \\
% DCNet~\cite{kim:2021:AAAI}  & \multicolumn{1}{|c}{23.95}     &  \multicolumn{1}{c|}{47.30}   & \multicolumn{1}{c}{28.95}           & \multicolumn{1}{c|}{56.07} &  \multicolumn{1}{c}{30.44}   &  \multicolumn{1}{c|}{58.29}  & \multicolumn{1}{c}{40.83}    \\
MAAF~\cite{maaf}        & \multicolumn{1}{|c}{21.30}   & \multicolumn{1}{c|}{44.20}      & \multicolumn{1}{c}{23.80}     &  \multicolumn{1}{c|}{48.60}     &  \multicolumn{1}{c}{27.90}   &  \multicolumn{1}{c|}{53.60}    & \multicolumn{1}{c}{36.57}      \\
RTIC~\cite{shin2021rtic}  & \multicolumn{1}{|c}{22.03}      & \multicolumn{1}{c|}{45.29}           & \multicolumn{1}{c}{27.37}           &  \multicolumn{1}{c|}{52.95}   &  \multicolumn{1}{c}{27.33}  & \multicolumn{1}{c|}{53.60} & \multicolumn{1}{c}{38.10}    \\

\hline
% FiLM$^*$     & \multicolumn{1}{|c}{20.79±0.49}    & \multicolumn{1}{c|}{45.58±0.68}  & \multicolumn{1}{c}{26.06±1.09}  &  \multicolumn{1}{c|}{53.05±0.54}         & \multicolumn{1}{c}{25.58±0.78}             &  \multicolumn{1}{c|}{52.68±0.69}   & \multicolumn{1}{c}{37.29±0.71}    \\
% MRN$^*$      & \multicolumn{1}{|c}{21.32±0.50}      & \multicolumn{1}{c|}{44.82±0.73}           & \multicolumn{1}{c}{25.66±0.74}           &  \multicolumn{1}{c|}{52.01±0.97}     &  \multicolumn{1}{c}{25.94±1.00}   &  \multicolumn{1}{c|}{52.72±0.74}    & \multicolumn{1}{c}{37.08±0.78}  \\
TIRG$^*$     & \multicolumn{1}{|c}{21.38±0.54}  & \multicolumn{1}{c|}{46.28±0.78}   & \multicolumn{1}{c}{25.82±0.39}  &  \multicolumn{1}{c|}{53.21±0.33} &  \multicolumn{1}{c}{26.73±0.72}   &  \multicolumn{1}{c|}{53.17±0.29}  & \multicolumn{1}{c}{37.77±0.21}  \\
MAAF$^*$  & \multicolumn{1}{|c}{23.55±0.31}   &  \multicolumn{1}{c|}{46.38±1.34}   & \multicolumn{1}{c}{28.75±0.63}  & \multicolumn{1}{c|}{54.48±0.49}   &  \multicolumn{1}{c}{29.70±0.45}   &  \multicolumn{1}{c|}{55.84±0.87}   & \multicolumn{1}{c}{39.78±0.68}  \\
RTIC$^*$       & \multicolumn{1}{|c}{23.03±0.63}& \multicolumn{1}{c|}{46.68±0.52}  & \multicolumn{1}{c}{26.86±0.74}  &  \multicolumn{1}{c|}{52.80±0.61} &  \multicolumn{1}{c}{27.21±0.89}   &  \multicolumn{1}{c|}{53.24±0.66} & \multicolumn{1}{c}{38.31±0.67}\\
\hline
 AACL &\multicolumn{1}{|c}{\textbf{24.82±0.62}} & \multicolumn{1}{c|}{\textbf{48.85±0.77}}  & \multicolumn{1}{c}{\textbf{29.89±0.65}}  & \multicolumn{1}{c|}{\textbf{55.85±0.87}}   & \multicolumn{1}{c}{\textbf{30.88±1.2}}   &  \multicolumn{1}{c|}{\textbf{56.85±1.16}} &\multicolumn{1}{c}{\textbf{41.19±0.88}}   \\

\end{tabular}}
\vspace{-5pt}
\caption{Comparison of image search with text feedback on FashionIQ. Averaged R@10/50 computed over all three categories. * denotes results obtained with the same image encoder and text encoder as AACL.}
\label{tab:fashionIQ}
\end{table*}

\begin{figure}[!t]
  \centering
  \includegraphics[width=0.7\linewidth]{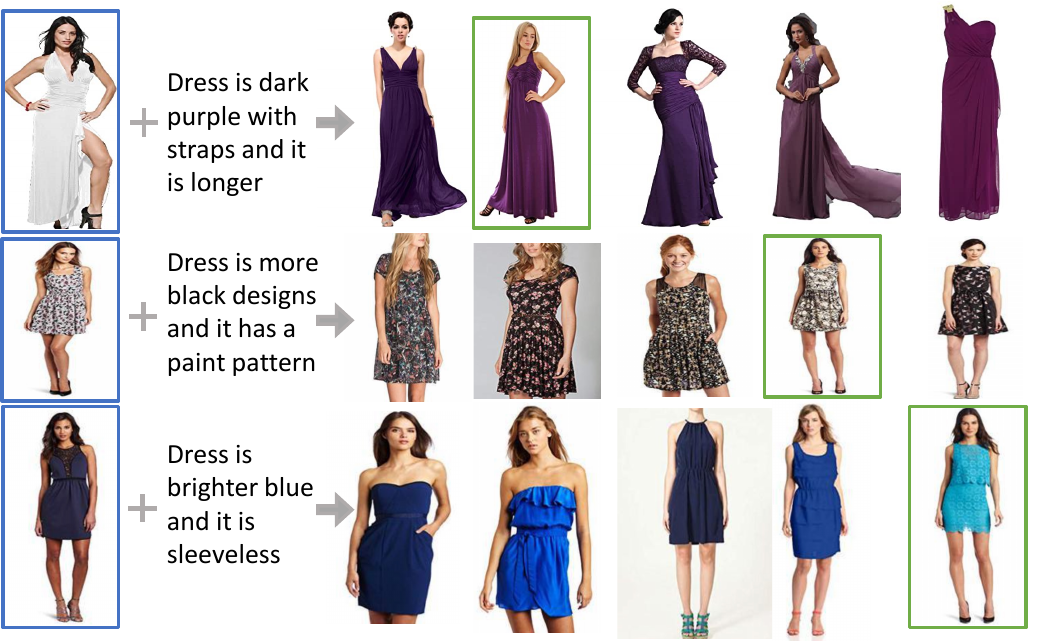}
  \vspace{-10pt}
  \caption{Qualitative results of AACL on FashionIQ dataset. Blue and green box indicate query and target images, respectively.}
  \label{fig:fashioniq}
\end{figure}

\noindent\textbf{Compared Methods:}
We compare the results of AACL with several methods, namely: FiLM, MRN, TIRG, ComposeAE, MAAF and RTIC. We explained them briefly in Section~\ref{sec:relatedwork}.

\subsection{FashionIQ} \label{fashionIQ}
FashionIQ~\cite{fashioniq} is a natural language based interactive fashion product retrieval dataset. It contains 77,684 images crawled from Amazon.com, covering three categories: Dresses, Tops\&Tees and Shirts. Among the 46,609 training images, there are 18,000 image pairs. Each pair is accompanied by on average two natural language sentences that describe one or multiple visual properties to modify in the reference image, such as \textit{“is shiny”} or \textit{``is blue in color and floral, and with white base"}. We follow the same evaluation protocol as~\cite{fashioniq}, using the same training split and evaluating on the validation set.
% Following the same evaluation protocol of composing image and text for retrieval~\cite{fashioniq} [A LITTLE CONFUSING], we use the same training split and evaluate on the validation set.
We report results on individual categories, as well as the average results over all three.

%%%FashionIQ~\cite{fashioniq} is a natural language based interactive fashion product retrieval dataset. It contains 77,684 images crawled from Amazon.com, covering three categories: Dresses, Tops\&Tees and Shirts. Among the 46,609 training images, there are 18,000 image pairs. Each pair is accompanied by on average two natural language sentences that describe one or multiple visual properties to modify in the reference image, such as \textit{“is shiny”} or \textit{``is blue in color and floral, and with white base"}.
%%%We follow the same evaluation protocol as~\cite{fashioniq}, using the same training split and evaluating on the validation set.
% Following the same evaluation protocol of composing image and text for retrieval~\cite{fashioniq} [A LITTLE CONFUSING], we use the same training split and evaluate on the validation set.
%%%We report results on individual categories, as well as the averaged results over the three categories.

Table~\ref{tab:fashionIQ} compares the performance of AACL and the other methods on FashionIQ. We observe that AACL is superior to all reported results by a large margin (top half). AACL even outperforms methods that include factors other than the composition module itself, such as the target image captions, model ensembles, and additional joint loss functions~\cite{composeAE}. We further note that AACL is actually complementary to some of these methods and could, in fact, be used as their composition modules. %[CHECK THIS STATEMENT]
For a like-to-like fair comparison, we also reproduced the best competitors, focusing on just the composition module itself. That is, we utilized the same image and text encoders---namely, Swin Transformer and DistilBERT---and the same optimizer. In this scenario AACL surpasses TIRG, RTIC, and MAAF by an overall margin of $3.42\%$, $2.88\%$ and $1.41\%$ respectively in average R@10 and R@50 scores.
Figure~\ref{fig:fashioniq} presents our qualitative results on FashionIQ. These results demonstrate that our model can handle complex and realistic text descriptions. We also observe that our model can jointly comprehend global appearance (e.g., colors, material), as well as local fine-grained details (e.g., straps and neckline, length of sleeves), for image search.

\renewcommand{\tabcolsep}{10pt}

\begin{table}[t]
\centering
\resizebox{0.7\linewidth}{!}{
\begin{tabular}{l|lll}
Model        & \multicolumn{1}{c}{R@1}   & \multicolumn{1}{c}{R@10}  & \multicolumn{1}{c}{R@50} \\ 
\toprule
% FiLM~\cite{perez2018film}          & 12.9±0.7                  & 39.5±2.1                  & 61.9±1.9                 \\
FiLM~\cite{perez2018film}          &\multicolumn{1}{c}{12.9}                  & \multicolumn{1}{c}{39.5}                  & \multicolumn{1}{c}{61.9} \\
MRN~\cite{kim2016multimodal}           & \multicolumn{1}{c}{13.4}           & \multicolumn{1}{c}{40.0}           & \multicolumn{1}{c}{61.9}                     \\
% TIRG~\cite{tirg}          & 14.1±0.6                  & 42.5±0.7                  & 63.8±0.8                 \\
TIRG~\cite{tirg}          & \multicolumn{1}{c}{14.1}                  & \multicolumn{1}{c}{42.5}                 & \multicolumn{1}{c}{63.8}                \\

ComposeAE~\cite{composeAE}     & \multicolumn{1}{c}{16.5} & \multicolumn{1}{c}{45.4} & \multicolumn{1}{c}{63.1} \\
% VAL~\cite{chen2020image}     &  \multicolumn{1}{c}{21.2}                     & \multicolumn{1}{c}{49.0}                      & \multicolumn{1}{c}{68.8}                     \\
% SAC~\cite{j2020sac}           &                           &                           &                          \\
DCNet~\cite{kim:2021:AAAI}         & \multicolumn{1}{c}{--}                        & \multicolumn{1}{c}{46.9}                     & \multicolumn{1}{c}{67.6}                    \\
MAAF~\cite{maaf}          & \multicolumn{1}{c}{18.94} & \multicolumn{1}{c}{--}            &  \multicolumn{1}{c}{--}      \\
% RTIC~\cite{shin2021rtic}          & --                        & --                        & --                       \\ 
\hline
% FiLM$^*$          &\multicolumn{1}{c}{17.23±0.70}       & \multicolumn{1}{c}{52.64±0.45}     & \multicolumn{1}{c}{72.95±0.91}                          \\
% MRN$^*$         &\multicolumn{1}{c}{17.65±0.26}      & \multicolumn{1}{c}{52.41±0.55}                           & \multicolumn{1}{c}{72.18±1.02}                          \\
TIRG$^*$          & \multicolumn{1}{c}{17.22±0.39}           &  \multicolumn{1}{c}{56.52±1.85}                   &  \multicolumn{1}{c}{75.60±0.09}                        \\
% ComposeAE$^*$     &                           &                           &                          \\
MAAF$^*$           &\multicolumn{1}{c}{17.79±0.98}      & \multicolumn{1}{c}{57.57±0.98}        &  \multicolumn{1}{c}{77.51±0.63}                         \\
RTIC$^*$          & \multicolumn{1}{c}{17.05±0.96}      & \multicolumn{1}{c}{54.65±0.79}        &  \multicolumn{1}{c}{75.54±1.63}   \\ 
\hline
AACL              &\multicolumn{1}{c}{\textbf{19.64±1.66}}      & \multicolumn{1}{c}{\textbf{58.85±1.01}}        &  \multicolumn{1}{c}{\textbf{78.86±0.43}}  \\
%  AACL$_{alter}$ \\
% AACL  &\multicolumn{1}{c}{±}      & \multicolumn{1}{c}{±}        &  \multicolumn{1}{c}{±}  \\
% AACL      & \multicolumn{1}{c}{19.27±0.69}      & \multicolumn{1}{c}{54.97±0.42}        &  \multicolumn{1}{c}{74.32±0.90}         \\ 

\end{tabular}}
\caption{Comparison of image search with text feedback on Fashion200k dataset.  * denotes our implementation results obtained with the same image encoder and text encoder as AACL.}
\label{tab:fashion200k}
\end{table}
\begin{figure}[t]
  \centering
  \includegraphics[width=0.68\linewidth]{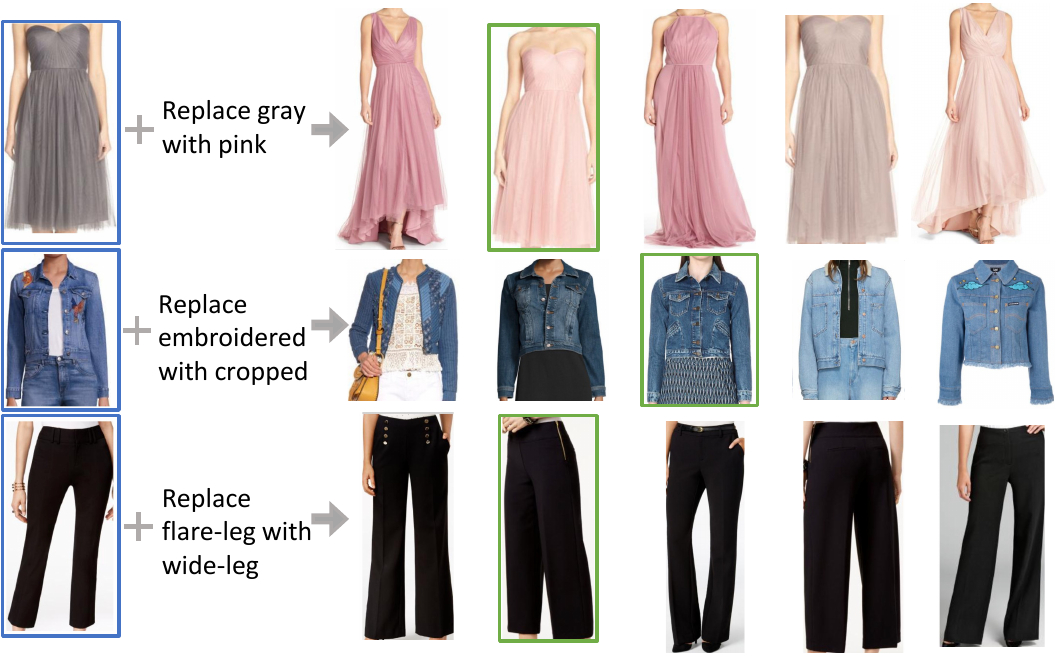}
      \vspace{-10pt}
  \caption{Qualitative results of AACL on Fashion200k dataset. Blue and green box indicate query and target images, respectively.}
  \label{fig:fashion200k}
\end{figure}

\vspace{-10pt}
\subsection{Fashion200k} \label{fashion200k}
Fashion200k~\cite{fashion200k} is a large-scale fashion dataset crawled from multiple online shopping websites. It contains more than 200k fashion images collected for attribute-based product retrieval. It also covers a diverse range of fashion concepts, with a total vocabulary size of 5,590. Each image is tagged with descriptive text corresponding to a product description, such as \textit{``beige v-neck bell-sleeve top''}. Following~\cite{tirg}, we use the training split of 172,049 images for training and the test set of 33,480 test queries for evaluation. During training, pairwise images with attribute-like modification texts are generated by comparing their product descriptions on-the-fly, e.g., \textit{``replace black with blue''} or \textit{``replace mini with midi''}.

% \begin{figure*}[!t]
% % \setlength{\belowcaptionskip}{-0.3cm} 
%      \centering
%      \begin{subfigure}{0.7\linewidth}
%     %  \setlength{\abovecaptionskip}{-0.02cm}
%          \centering
%           \includegraphics[width=.9\linewidth]{latex/figures/fashioniq.pdf}
%           \captionof{figure}{FashionIQ dataset.}
%           \label{fig:fashioniq}
%      \end{subfigure}
%     \hfill
%      \begin{subfigure}{0.7\linewidth}
%     %  \setlength{\abovecaptionskip}{-0.02cm}
%          \centering
%           \includegraphics[width=.9\linewidth]{latex/figures/fashion200k.pdf}
%           \captionof{figure}{Fashion200k dataset.}
%           \label{fig:fashion200k}
%      \end{subfigure}
%      \hfill
%     \caption{Qualitative results of AACL on (a) FashionIQ and (b) Fashion200k dataset. Blue and green box indicate query and target images, respectively.
%     }
%     \label{fig:}
% \end{figure*}

\begin{figure}[t]
  \centering
  \setlength{\abovecaptionskip}{-0.02cm}
    \setlength{\belowcaptionskip}{-3cm} 
   \includegraphics[width=0.8\linewidth]{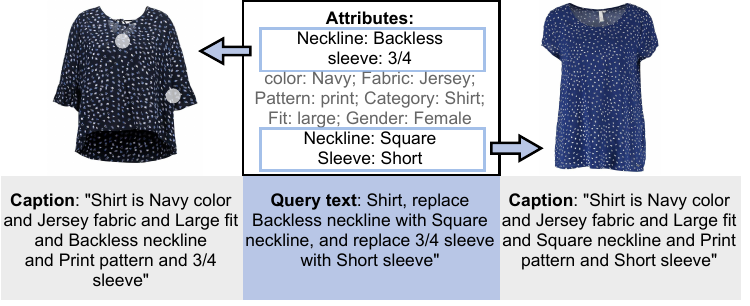}
   \caption{Example of image pair and generated text query from Shopping100k dataset. Gray words indicate shared attributes.}
   \label{fig:shopping100k}
\end{figure}

% Please add the following required packages to your document preamble:
% \usepackage[table,xcdraw]{xcolor}
% If you use beamer only pass "xcolor=table" option, i.e. \documentclass[xcolor=table]{beamer}

\renewcommand{\tabcolsep}{1pt}
\begin{table}[t]
\setlength{\abovecaptionskip}{-2pt}
\setlength{\belowcaptionskip}{-10cm} 
\centering
\resizebox{0.9\linewidth}{!}{
\begin{tabular}{cccccccccccc}
\hline
& Jacket   & Shirt  & T-shirt  & Jumper   & Shorts  & Trouser & Jean   & Swim    &  Bottoms\tablefootnote{Full name of category ``Bottoms" is ``Tracksuit Bottoms".}  & Skirt & Dress  \\ 
\hline
  &7,528     & 14,853 & 22,071   & 11,797   & 5,099   & 4,630   & 6,229  & 5,497     &3,726            &2,528   &12,119 \\
\hline
\end{tabular}}
\caption{Number of images in select categories (count $>$ 2k) in Shopping100k dataset. }
\label{tab:shopping100k_cat}
\end{table}
% Coat   &   \\
% Jacket & \\
% Suit   &  \\
% Shirt  & \\
% T-shirt &22,071 \\
% Jumper & 11,797 \\
% Shorts & 5,099 \\
% Trouser & 4,630 \\
% Jean & 6,229 \\
% Swimming & 5,497 \\
% Jumpsuit & 1,287 \\
% Pyjamas & 1,091 \\
% Tracksuit Bottoms &3,726 \\
% Tracksuit 218 \\
% Skirt 2,528 \\
% Dress 12,119

%%%Fashion200k~\cite{fashion200k} is a large-scale fashion dataset crawled from multiple online shopping websites. It contains more than 200k fashion images collected for attribute-based product retrieval. It also covers a diverse range of fashion concepts, with a total vocabulary size of 5,590. Each image is tagged with descriptive text as product description, such as \textit{“beige v-neck bell-sleeve top”}. Following~\cite{tirg}, we use the training split of 172,049 images for training and the test set of 33,480 test queries for evaluation. During training, pairwise images with attribute-like modification texts are generated by comparing their product descriptions on-the-fly, i.e., \textit{"replace black to blue"},  \textit{"replace mini with midi"}.

Table~\ref{tab:fashion200k} shows our model achieves compelling results compared to other methods, most notably for R@1 where AACL outperforms the best competitor MAAF by a relative margin of $9.4\%$. We also observe that token based methods, namely MAAF and AACL, perform better than residual based methods. This indicates that the rich information contained in tokens is beneficial for feature composition. Figure~\ref{fig:fashion200k} shows our qualitative results on Fashion200k. Our model is able to retrieve new images that resemble the reference image, while changing certain attributes conditioned on text feedback---e.g., fit, color and length. We also observe that all retrieved images share the same semantics and are visually similar to the target image, indicating the quantitative performance is potentially underestimated. 

%%%Table~\ref{tab:fashion200k} shows our model demonstrates compelling results compared to all other alternatives, e.g. the relative improvement of R@1 
%%%AACL outperforms the 
%%%best competitor MAAF with an improved margin of $1.85\%$, $1.28\%$, $1.35\%$ in R@1, R@10 and R@50 respectively. We also observe that token based methods, i.e., MAAF and AACL, perform better than residual based methods. This indicates that the rich information contained in tokens is beneficial for feature composition.
%%%Figure~\ref{fig:fashion200k} shows our qualitative results on Fashion200k. We observe our model is able to retrieve new images that resemble the reference image, while changing certain attributes conditioned on text feedback---e.g., fit, color and length. We can also observe that all retrieved images share the same semantics and are visually similar to the target image, which indicates the quantitative performance is underestimated. 

\vspace{-10pt}
\subsection{Shopping100k} \label{shopping100k}
Shopping100k~\cite{shop100k} is a large-scale fashion dataset of individual clothing items extracted from different e-commence providers. It contains 101,021 images of 12 fashion attributes, covering the following categories: ``collar", ``color", ``fabric", ``fastening", ``fit", ``gender", ``length", ``neckline", ``pattern", ``pocket", ``sleeve length", and ``sport". A total of 151 different labels are generated by combinations of different attributes and the corresponding attributes values. Compared to FashionIQ and Fashion200k, the Shopping100k dataset is more diverse and only contains garments in isolation. In addition, FashionIQ and Fashion200k only contain 3 and 5 apparel categories, respectively.  

\begin{table*}[t]
\resizebox{1.0\linewidth}{!}{
\begin{tabular}{l|c|c|c|c|c|c|c|c|c|c|c|c}
% Methods & \multicolumn{12}{c}{Categories} \\ 
% \hline
 Model & Dress &Jacket	&Jean	&Jumper	&Shirt	&Shorts	&Skirt	&Swimming	&T-shirt	&Bottoms	&Trouser & Average \\ \toprule
\multicolumn{1}{c}{Recall@1} \\
\hline

TIRG$^*$  &6.81±0.58	&10.46±0.97	&4.83±1.43	&11.87±1.26	&13.15±1.25	&12.38±1.16	&10.92±1.22	&13.51±1.49	&11.87±0.80	&8.32±0.60	&13.03±1.77	&10.65±0.37  \\		
MAAF$^*$  &  7.05±0.86	&12.43±0.76	&5.79±1.34 &{13.19±0.88}	&\textbf{14.44±1.28}	&13.21±1.68	&12.11±0.77	&12.41±0.71	&12.89±1.16	&\textbf{10.28±1.35	}&12.89±0.87 &11.52±0.39\\
RTIC$^*$  &  6.80±0.09	&11.70±0.90	&5.27±0.90	&12.08±1.39	&13.93±1.33	&11.83±0.97	&10.96±1.44	&13.18±0.99	&12.60±0.99	&8.49±0.65	&11.70±1.70 &10.78±0.44\\
AACL        &\textbf{7.70±0.67}	&\textbf{12.63±0.93}	&\textbf{7.27±0.96}	&\textbf{13.30±0.31}	&14.21±0.52	&\textbf{14.38±1.14}	&\textbf{14.55±1.22}	&\textbf{16.22±1.02}	&\textbf{13.66±0.28}	&10.00±0.53	&\textbf{14.14±0.63}  &\textbf{12.55±0.32}\\
\hline
\multicolumn{1}{c}{Recall@10}  \\
% & Jacket & Shirt & T-shirt & Jumper & Shorts & Trouser & Jean & Swim & Tracksuit & Skirt & Dress & Average \\ 
\hline
% FiLM$^*$  & 32.70±1.68	&51.15±0.58	&29.06±0.08	&51.31±0.06	&51.82±0.33	&49.72±0.10	&54.26±1.00	&53.01±0.48	&48.42±0.57	&{42.59±0.33}	&54.91±1.75	&47.54±0.60\\
% MRN$^*$  &32.78±0.71	&48.56±0.69	&28.32±1.49	&50.66±0.64	&51.88±0.25	&51.85±1.42	&57.19±1.50	&52.40±0.97	&47.24±1.16	&43.32±0.67	&53.06±0.06	&46.84±0.49 \\
TIRG$^*$  &34.22±0.53	&49.86±0.47	&29.23±0.48	&51.08±0.89	&50.22±0.72	&50.43±0.52	&55.85±0.58	&51.86±1.49	&47.19±1.04	&41.69±0.59	&51.06±1.28	&46.61±0.35 \\
MAAF$^*$  &{35.01±1.85}	&{51.48±1.67}	&\textbf{31.78±1.12}	&{51.70±2.45}	&{52.15±1.96}	&50.64±1.30	&54.70±3.36	&54.74±2.46	&\textbf{49.31±1.79}	&44.00±2.87	&52.08±0.63  &47.96±0.65\\
RTIC$^*$  &	33.17±1.92	&50.51±2.11	&29.21±4.36	&48.92±3.39	&50.90±2.89	&50.29±0.74	&51.96±2.09	&51.62±2.02	&46.71±2.41	&42.24±1.31	&51.46±1.25	&46.09±1.03	\\
AACL     & \textbf{35.16±0.54}	&\textbf{51.63±1.33}	&30.80±1.79	&\textbf{52.31±0.89}	&\textbf{52.52±1.32}	&\textbf{54.63±1.66}	&\textbf{57.54±0.95}	&\textbf{56.13±2.13}	&{49.18±1.40}	&\textbf{46.69±1.06}	&\textbf{54.63±1.72} &\textbf{49.20±0.46}\\
\hline
\multicolumn{1}{c}{Recall@50} \\
% & Jacket & Shirt & T-shirt & Jumper & Shorts & Trouser & Jean & Swim & Tracksuit & Skirt & Dress & Average \\ 
\hline
% FiLM$^*$  &68.11±0.85	&81.12±0.41	&64.52±4.48	&81.19±0.17	&83.06±0.16	&80.73±0.19	&88.10±0.01	&85.77±0.48	&79.96±0.14	&80.21±0.78	&86.96±1.32	&80.16±1.28 \\
% MRN$^*$  &67.03±0.10	&82.54±0.30	&64.27±7.13	&81.16±0.82	&82.77±0.18	&81.53±0.57	&87.40±2.66	&84.34±0.38	&79.75±0.33	&78.21±2.02	&85.94±0.37	&79.54±2.08 \\
TIRG$^*$  &66.15± 0.80	&81.50±0.38	&62.47±0.19	&80.74±2.40	&82.43±0.28	&81.36±0.95	&85.57±1.66	&83.91±1.20	&79.32±1.81	&77.94±1.18	&85.02±1.35	&78.76± 0.69 \\
MAAF$^*$  & 68.42±1.42	&82.73±2.29	&63.24±2.94	&{82.28±1.36}	&84.41±1.90	&82.06±1.66	&{88.19±0.78}	&\textbf{85.32±2.27}	&\textbf{81.07±1.34}	&81.17±0.67	&86.75±0.82	&80.51±0.56 \\
RTIC$^*$  &67.30±2.12	&81.92±	2.42	&\textbf{64.30±5.31}	&80.27±2.37	&83.45±1.58	&82.22±1.88	&84.71±1.57	&84.15±2.46	&78.87±1.95	&79.47±0.88	&85.37±1.92	&79.27±1.12 \\
AACL & \textbf{69.21±0.37}	&\textbf{83.30±1.77}	&{63.92±3.59}	&\textbf{82.30±0.36}	&\textbf{84.75±1.21}	&\textbf{85.50±1.30}	&\textbf{88.94±0.78}	&{85.31±1.52}	&{80.54±1.18}	&\textbf{82.83±0.88}	&\textbf{87.61±0.76}	&\textbf{81.29±1.11}\\

\end{tabular}}
\caption{Comparison of image search with text feedback on our modified Shopping100k dataset. Averages are computed over all categories. * denotes our implementation results obtained with the same image encoder and text encoder as AACL.}
\label{tab:shopping100k}
\end{table*}

Each image in Shopping100k is tagged with the attributes and attribute values, such as \textit{``Neckline: Backless, Sleeve: 3/4, Color: Navy, Fabric: Jersey, Pattern: Print, Category: Shirt, Fit: Large, Gender: Female''}.
There are 15 high-level apparel categories. To generate the dataset for image retrieval with text feedback, we remove categories that contain fewer than 2,000 images, namely ``coat", ``suit", ``jumpsuit", ``pyjamas", and ``tracksuit". The final set of 11 categories is listed in Table~\ref{tab:shopping100k_cat} along with the number of images in each category. A training split with 76,867 images and a validation split with 19,210 images is randomly sampled from these remaining categories.
%%%Each image in Shopping100k is tagged with the attributes and attribute values, such as \textit{"category: T-shirt , fit: small, neckline: off-the-shoulder, pattern: plain, sleeve: sleeveless, gender: female"}.
%%%There are 15 high-level apparel categories. To generate the dataset for image retrieval with text feedback, we remove categories that contain fewer than 2,000 images, namely ``coat", ``suit", ``jumpsuit", ``pyjamas" and ``tracksuit". The final list of categories is presented in Table~\ref{tab:shopping100k_cat} along with the number of images in each category. 
%%%A training split with 76,867 images and a validation split with 19,210 images is randomly sampled from these remaining categories.

To generate the training image pairs and modification text, we first derive a descriptive caption for each image using its tagged attribute values by concatenating the category with ``is", followed by attributes joined by ``and"---e.g., \textit{``Shirt is Navy color and Jersey fabric and Large fit and Backless neckline and Print pattern and 3/4 sleeve"}. Queries are created by selecting image pairs that differ in two attributes in the description. Note that we constrain the image pairs to be from the same apparel category and gender. The modification text is created with the apparel category plus the attribute modifications following the pattern ``replace xx with xx"---i.e. \textit{``Shirt, replace Backless neckline with Square neckline, and replace 3/4 sleeve with Short sleeve."} (Figure~\ref{fig:shopping100k}). During training, the query and target image pairs are selected on-the-fly based on the number of attributes we specify. For our experiments, 16,237 fixed test query pairs are generated from the validation set for performance evaluation.

%%%To generate the training image pairs and modification text, we first derive a descriptive caption for each image using its tagged attribute values by concatenating the category with `is', followed by attributes joined by `and'---e.g.,  \textit{``T-shirt is small fit and off-the-shoulder neckline and plain pattern and sleeveless sleeve"}. Queries are created by selecting image pairs that differ in a small number of attributes (either 1 or 2) in the description. Note that we constrain the image pairs to be from the same apparel category and gender. The modification text is created with the apparel category plus the attribute modifications following the pattern ``replace xx with xx"---i.e. \textit{``T-shirt, replace Small fit with Regular fit, and replace Sleeveless sleeve with Elbow sleeve."} (Figure~\ref{fig:shopping100k}) [TEXT EXAMPLE DOESN'T MATCH FIGURE]. During training, the query and target image pairs are selected on-the-fly based on the number of attributes we specify. For the experiments with 2-attribute modifications, 16,237 fixed test query pairs are generated from the validation set for performance evaluation.
\begin{figure*}[t]
     \centering
     \begin{subfigure}{0.75\linewidth}
         \centering
         \includegraphics[width=\linewidth]{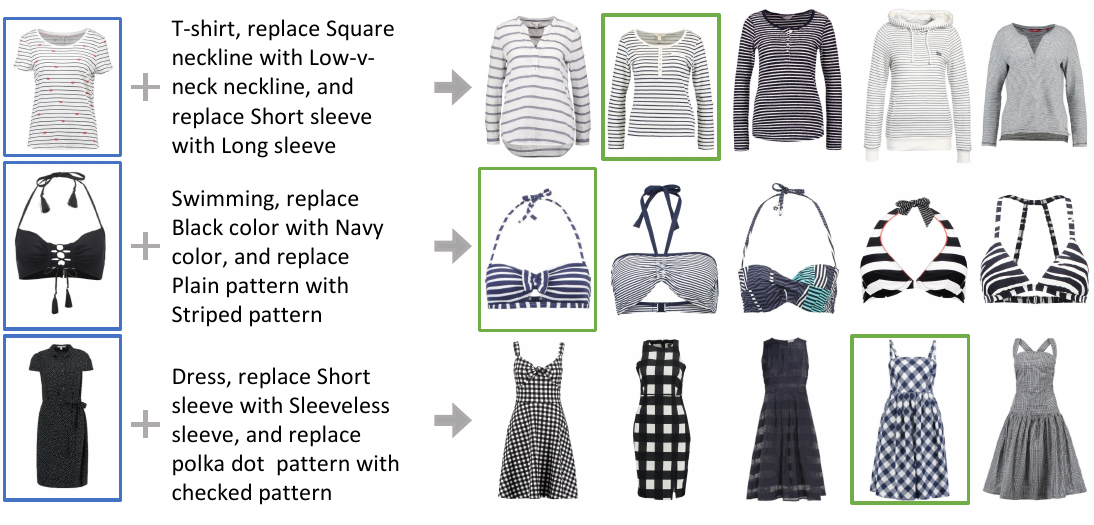}
         \caption{Female examples}
         \label{female}
     \end{subfigure}
    \hfill
     \begin{subfigure}{0.75\linewidth}
         \centering
         \includegraphics[width=\linewidth]{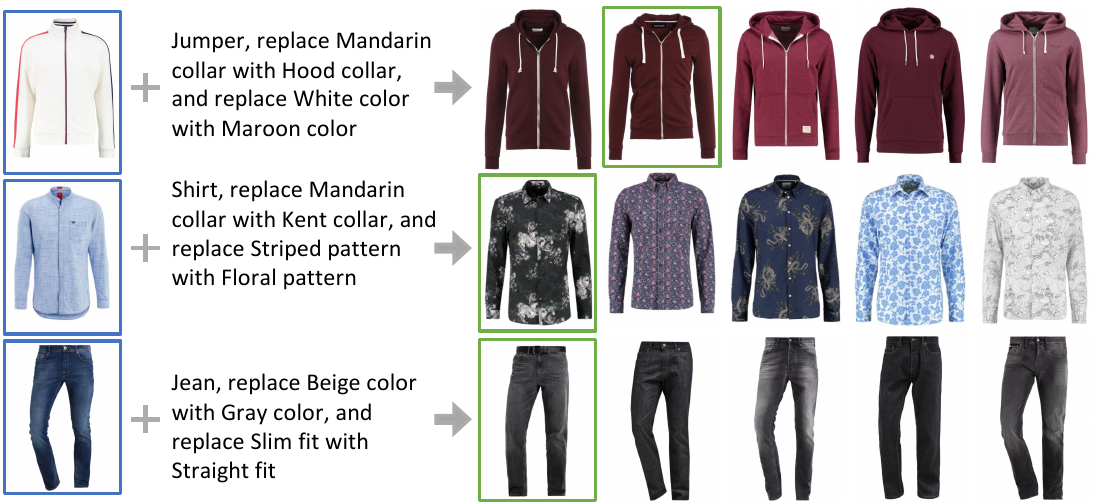}
         \caption{Male examples}
         \label{male}
     \end{subfigure}
     \hfill
    \caption{Qualitative results of AACL on (a) Female and (b) Male set of our modified Shopping100k dataset. Blue and green box indicate query and target images, respectively.
    }
    \label{fig:shop100k_quality}
\end{figure*}
Table~\ref{tab:shopping100k} compares our approach to other methods on Shopping100k. Our model is shown to clearly outperform the SOTA baselines. Figure~\ref{fig:shop100k_quality} presents some qualitative examples. These examples yield three observations. First, our model is capable of understanding rich image-text representations, including global attributes such as color, pattern, and fit, as well as local attributes such as collar, neckline, and sleeves. Second, our model is capable of using the text information to selectively modify the query images. As an example, for the first query the retrieved images preserve the striped pattern even though it is not requested in the text feedback. Five of the top-5 retrieved candidates fulfill the ``long sleeves" requirement and four candidates have ``low-v-neck". Third, the model is capable of capturing minor modifications such as ``kent collar" \textit{vs.} ``mandarin collar", suggesting it can be successfully utilized in fine-grained search. 

%%%Table~\ref{tab:shopping100k} shows our comparison with existing methods when there are 2 different attributes for each pair. Our model demonstrates the clear superiority to the SOTA baselines. 
%%%Figure~\ref{fig:shop100k_quality} presents some qualitative examples on the Shopping100k dataset. These examples yield three observations. First, our model is capable of understanding the rich image-text representations, including global attributes such as color, pattern, and fit, as well as the local attributes such as collar, neckline, and sleeves. Second, our model is capable of using the text information to selectively modify the query images. For the first query, the retrieved images still preserve the striped pattern even though it is not requested in the text feedback. Five of the top-5 retrieved candidates fulfill the ``long sleeves" requirement and four candidates have ``low-v-neck". Third, the model is capable of capturing minor modifications such as ``kent collar" \vs ``mandarin collar", suggesting it can be successfully utilized in fine-grained search. 

\vspace{-10pt}
\subsection{Ablation Study}

\begin{figure}[t]
\begin{floatrow}
\capbtabbox{%
\renewcommand{\tabcolsep}{7pt}
\resizebox{0.94\linewidth}{!}{\begin{tabular}{p{8em} |l|cc}
            Stage(s) & Recall@10 &Recall@50 \\ \toprule
            Stage 2 + 3 + 4  & 48.78 & 80.74\\
            Stage 3 + 4   & \textbf{49.20}  & \textbf{81.29}\\
            Stage 4     & 48.56  & 81.25\\
        \end{tabular}}

}{%
  \caption{Ablation of using tokens from different Swin Transformer stages on our modified Shopping100k dataset.}%
  \label{tab:shop100kstage}
}
\ffigbox{%
\includegraphics[width=\linewidth]{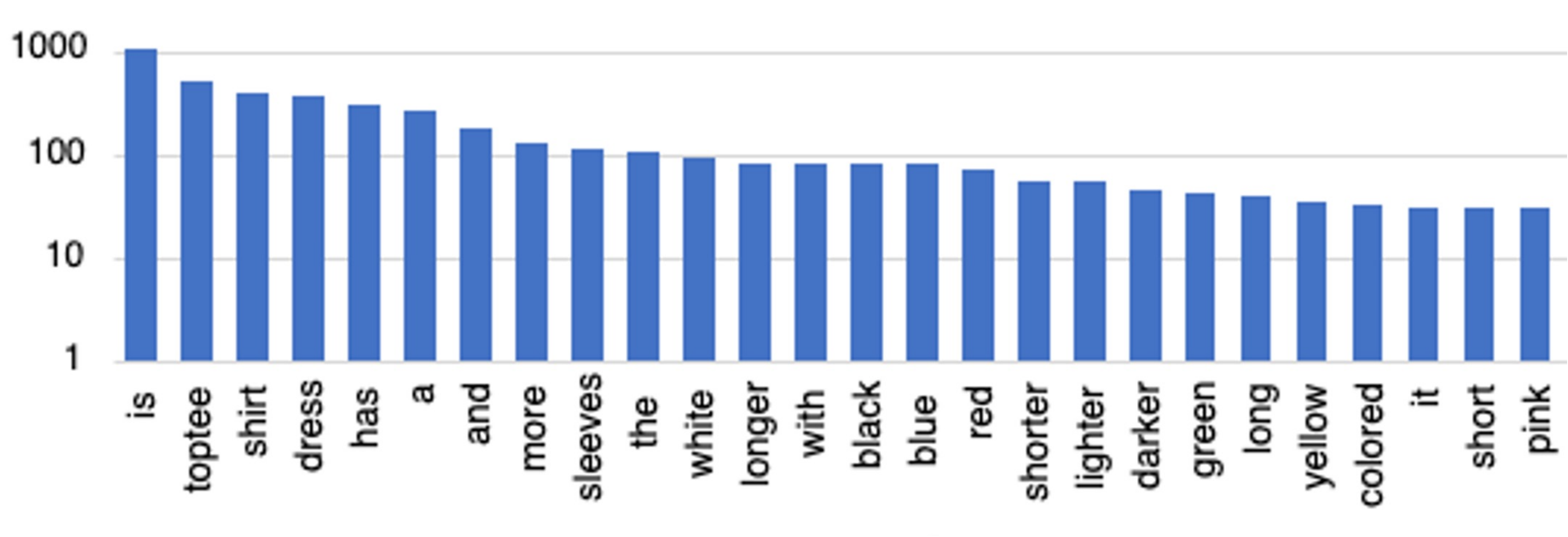}
}{%
   \caption{Frequency plot of words in FashionIQ validation set assigned high attention value by the AACL model.}%
   \label{fig:words}
}
\end{floatrow}
\end{figure}

\noindent\textbf{Image representation:}
Table~\ref{tab:shop100kstage} compares the performance of AACL when using different image representations from the Swin Transformer on our modified Shopping100k dataset. The experiments reveal that using image tokens from Stages 3 and 4 is most effective for this task. The concatenation of two stages from the encoder considers richer forms of image representation. Somewhat surprisingly, concatenating representations from Stage 2 does not seem to benefit the task. This may suggest that at some point, the lower level information may distract the model from capturing meaningful global contextual information. 

\renewcommand{\tabcolsep}{7pt}

\begin{table*}[ht]
\centering
\setlength{\belowcaptionskip}{-0.20cm} 
\resizebox{\linewidth}{!}{
\begin{tabular}{l|c|c|c|c|c|c|c|c|c|c|c|c}
 Dataset & Dress &Jacket	&Jean	&Jumper	&Shirt	&Shorts	&Skirt	&Swimming	&T-shirt	&Bottoms	&Trouser & Average \\ \toprule

\multicolumn{1}{l}{Recall@10} \\
\hline
1 attribute &\underline{30.11}	&52.48	&47.71	&53.34	&51.36	&55.39	&\underline{52.17}	&\underline{54.29}	&\underline{48.90}	&47.69	&50.24	&49.42 \\
1 or 2 attributes &33.64    	&54.25	&46.28	&{54.38}	&\underline{51.11}        &	{54.36}            	&55.25	            &{58.98}	&{52.09}	&{46.44}	&\underline{49.43}	        &{50.56}\\
2 attributes & {35.16}	&\underline{51.63}	&\underline{30.80}	&\underline{52.31}	&52.52	&\underline{54.63}	&{57.54}	&{56.13}	&49.18	&\underline{46.69}	&{54.63} &\underline{49.20}

\\

\hline
\multicolumn{1}{l}{Recall@50} \\
\hline
1 attribute   &\underline{61.52}	   &84.78	&80.09	&84.02	&84.10	&86.83	&89.02	&\underline{84.70}	&81.04	&\underline{80.03}	&\underline{84.66}	&81.89\\
1 or 2 attributes&66.78   &{85.16} &{81.48}	&{83.38}	&\underline{82.72} &{86.25}	&{89.62}	&{87.42}	&81.67 &83.07	&86.81	&{83.12}\\
2 attributes   &{69.21}	&\underline{83.30}	&\underline{63.92}	&\underline{82.30}	&{84.75}	&\underline{85.50}	&\underline{88.94}	&{85.31}	&\underline{80.54}	&{82.83}	&{87.61}	&\underline{81.29}\\

\end{tabular}}
\caption{Comparison of AACL when constructing Shopping100k dataset with different number of attributes. Lowest value is underlined.}
\label{tab:ablation_att}
\end{table*}

\renewcommand{\tabcolsep}{14pt}

\begin{table}[ht]
\centering
\setlength{\abovecaptionskip}{-0.01cm} 
\setlength{\belowcaptionskip}{-0.20cm} 
\resizebox{0.5\linewidth}{!}{

\begin{tabular}{l|c|c}
 Method & Recall@10 &Recall@50 \\ \toprule
Additive$\rightarrow$Dot-Product 	&48.37  & 80.14\\
Product$\rightarrow$Addition 	&48.56 &80.45 \\
AACL  &\textbf{49.20} &\textbf{81.29}\\
\end{tabular}

}
\caption{Ablation of self-attention layer on our modified Shopping100k dataset. We separately examine substituting additive self-attention with standard dot-product and changing the Hadamard product to addition.}
\vspace{-11pt}
%Additive$\rightarrow$Dot-Product: substitute additive self-attention layer with standard dot-product. Product$\rightarrow$Addition: change Hadamard product to addition.}
\label{tab:ablation_shop100k}
\end{table}

%%%Table~\ref{tab:ablation} compares the performance of AACL when using different image representations from the Swin transformer. The experiments reveal that using image tokens from Stages 3 and 4 is most effective for this task. The concatenation of two stages from the image encoder considers richer forms of image representation. Somewhat surprisingly, concatenating image representations from Stage 2 does not seem to benefit the task. That may suggest that at some point the lower level information may distract the model from capturing meaningful global contextual information. 

\noindent\textbf{Number of attributes:} 
In Table~\ref{tab:ablation_att}, we see the effect of the number of attributes that differ on the Shopping100k dataset. We constrain the modification text to have varying numbers of differing attributes: 2 attributes, 1 or 2 attributes, or 1 attribute. Having 2 differing attributes is seen to be the most difficult case and so we choose it to compare with the other methods in Table~\ref{tab:shopping100k}. 

%%%We study the influence of using different numbers of attributes on the Shopping 100k dataset (see Table~\ref{tab:ablation_att}). We constrain the modification text ask for different number of attributes, namely 2 attributes, 1 or 2 attributes,or 1 attribute. Requiring 2 attributes inside one text feedback is the hardest setting and we choose the hardest setting to compare with other competitors in Table~\ref{tab:shopping100k}. 

% \begin{figure}[t]
% \setlength{\abovecaptionskip}{-0.1pt}
% \setlength{\belowcaptionskip}{-4pt}
%   \centering
%   \includegraphics[width=0.5\linewidth]{latex/figures/words.pdf}
%   \caption{Frequency plot of words in FashionIQ validation set assigned high attention value by the AACL model.}
%   \label{fig:words}
% \end{figure}

% \noindent\textbf{Additive self-attention:}
% We study substituting the additive self-attention layer with a standard dot-product one. The results are shown in Table~\ref{tab:ablation_interact}:  ``Additive $\rightarrow$ Dot-Product". The results show the effectiveness of our additive self-attention layer.

%%%\noindent\textbf{Influence of additive self-attention layer.}
%%%We study the influence of substituting the additive self-attention layer to standard dot-product attention layer and show the results in Table~\ref{tab:ablation_interact} Additive $\rightarrow$ Dot-Product.  
\noindent\textbf{Additive attention:}
To assess the importance of additive attention, we perform a comparison by substituting with dot-product attention. Table~\ref{tab:ablation_shop100k} ``Additive$\rightarrow$Dot-Product" shows the comparison on our modified Shopping100k dataset. From these results, we that AACL does benefit consistently from the additive attention. {In addition, dot product attention is more computationally expensive than additive attention ($O(n^2)$ \textit{vs.} $O(n)$) and as such the benefits of additive attention extend beyond evaluation performance gains.}

\noindent\textbf{Interaction function:}
We study the effect of using different functions, namely addition and Hadamard product, to model the interactions between the context vector and the individual tokens. We compare the standard AACL and this variant on Shopping100k. The results are shown in Table~\ref{tab:ablation_shop100k} ``Product$\rightarrow$Addition". The Hadamard product performs consistently better than addition, indicating this form  of non-linear modeling is beneficial.%. While addition only models the linear interactions between the context vector and individual tokens, the Hadamard product is capable of modeling non-linear relationships between two variables.

%%%\noindent\textbf{Influence of interaction functions.}
%%%We study the influence of using different functions, namely addition and Hadamard product, to model the interactions between the global context vector and individual tokens in AACL. We compare AACL and its variant on the FashionIQ dataset. The results are shown in Table~\ref{tab:ablation_interact} Product$\rightarrow$Addition. The Hadamard product works consistently better in fusing the features than addition. While addition only models the linear interactions between the context vector and individual tokens, the Hadamard product is capable of modeling non-linear relationships between two variables.
\vspace{-10pt}
\subsection{Additive Attention Visualization}
To interpret the attention learned by AACL, we count the number of instances of words with high normalized attention scores from the FashionIQ validation set. The word attention scores are normalized as follows: We first multiply the $\alpha_i$ in Equation~\ref{eq:alpha} across all blocks to get the total attention flow for each token. Subsequently, the minimum word token flow score is mapped to zero and the maximum to one. We apply a threshold of $0.8$ for high scores. The top-30 most important words are shown in Figure~\ref{fig:words}. The model focuses on relationship words such as ``is'' and  ``has'', which are important for learning the relationship of the attributes. The fact that our model pays so much attention to categories of clothing indicates it has learned to retrieve garments from the same category automatically. The model has also learned the importance of attributes such as ``sleeves'', ``white'', and ``longer'', for modifying the query image.

\begin{figure}[!t]
  \centering
\setlength{\abovecaptionskip}{-0.2cm}
\setlength{\belowcaptionskip}{-0.4cm}    \includegraphics[width=0.65\linewidth]{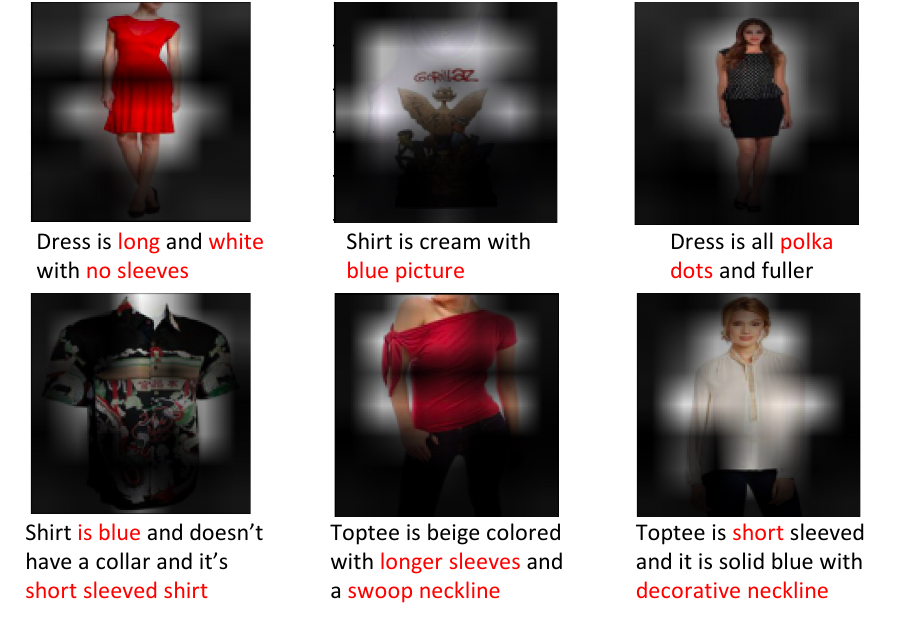}
   \caption{Attention visualization of AACL model on FashionIQ dataset. Words with high attention value in \textcolor{red}{red}.}
   \label{fig:visualization}
  \vspace{-10pt}
\end{figure}

To further interpret the attention learned by AACL, we visualize the attended regions in Figure \ref{fig:visualization}. We apply a mask based on the attention flow (as calculated above) to the input query image. Note that, since we are using the Swin Transformer as the image encoder, the encoded feature maps are 7 $\times$ 7 and the resulting visualization resolution appears lower than with other models. Nevertheless, we do
observe that the spatially attended regions vary with the query text. This indicates that the additive self-attention selects different visual content to transform conditioned on the text query.

%%%\noindent\textbf{Attention visualization.}
%%%To further interpret the attentions learnt by AACL, we visualize the attended regions in Figure \ref{fig:visualization}. We observe that the spatially attended regions vary with the query text. This indicates the additive attention stream selects different visual content to transform conditioned on text query. 
%----------------------------------------------------------------------------------------------------------------------

\vspace{-10pt}
\section{Conclusion}
\vspace{-10pt}

We present AACL, a novel and general-purpose solution to the challenging task of image search with text feedback. This framework features an additive self-attention layer that selectively preserves and transforms multi-level visual features conditioned on text semantics to derive an expressive composite representation. We validate the efficacy of AACL on three datasets, and demonstrate its consistent superiority in handling various text feedback for natural language expression. Overall, our work provides a novel approach along with a comprehensive evaluation, which collectively advance the research in interactive visual search using text feedback. %As a future extension, our approach also opens up the possibility to bridge multiple modalities for multi-modal retrieval in other domains such as home decor. 

%%%We present AACL, a novel and general-purpose approach to tackle the challenging task of image search with text feedback. This framework features an additive self-attention layer that selectively preserves and transforms multi-level visual features conditioned on semantics to derive an expressive composite representation. We validate the efficacy of AACL on three datasets, and demonstrate its consistent superiority in handling various text feedback for natural language expression. Overall, this work provides a novel approach along with a comprehensive evaluation, which collectively advance the research in interactive visual search using text feedback. As a future extension, our approach also opens up the possibility to bridge multiple modalities for multi-modal retrieval in other domains such as home decor.

\clearpage
% ---- Bibliography ----
%
% BibTeX users should specify bibliography style 'splncs04'.
% References will then be sorted and formatted in the correct style.
%
\bibliographystyle{splncs04}
\bibliography{egbib}
\end{document}